\documentclass[11pt,letterpaper]{article}
\usepackage{emnlp2017}
\usepackage{times}
\usepackage{latexsym}

\emnlpfinalcopy

\def\emnlppaperid{223}

\usepackage{graphicx}
\usepackage{amsmath}
\usepackage{amsfonts}
\usepackage{mathtools}
\usepackage{booktabs}
\usepackage{graphicx}
\usepackage{xcolor}
\usepackage{enumitem}

\DeclareMathOperator{\score}{s}
\DeclareMathOperator{\softmax}{softmax}
\DeclareMathOperator{\lstm}{LSTM}
\DeclareMathOperator{\latticelstm}{LatticeLSTM}
\title{Neural Lattice-to-Sequence Models for Uncertain Inputs}

\author{Matthias Sperber$^{1}$, Graham Neubig$^{2}$, Jan Niehues$^{1}$, Alex Waibel$^{1}$ \\
  $^{1}$Karlsruhe Institute of Technology, Germany \\
  $^{2}$Carnegie Mellon University, USA \\
  {\tt \small matthias.sperber@kit.edu} \\
}

\date{}

\begin{document}

\maketitle

\begin{abstract}
The input to a neural sequence-to-sequence model is often determined by an up-stream system, e.g.\ a word segmenter, part of speech tagger, or speech recognizer. These up-stream models are potentially error-prone. Representing inputs through word lattices allows making this uncertainty explicit by capturing alternative sequences and their posterior probabilities in a compact form.
In this work, we extend the TreeLSTM \cite{Tai2015} into a LatticeLSTM that is able to consume word lattices, and can be used as encoder in an attentional encoder-decoder model. We integrate lattice posterior scores into this architecture by extending the TreeLSTM's child-sum and forget gates and introducing a bias term into the attention mechanism. We experiment with speech translation lattices and report consistent improvements over baselines that translate either the 1-best hypothesis or the lattice without posterior scores.
\end{abstract}

\section{Introduction}

In many natural language processing tasks, we will require a down-stream system to consume the input of an up-stream system, such as word segmenters, part of speech taggers, or automatic speech recognizers. Among these, one of the most prototypical and widely used examples is speech translation, where a down-stream translation system must consume the output of an up-stream automatic speech recognition (ASR) system.

\begin{figure}[tb]
\includegraphics[width=7.5cm]{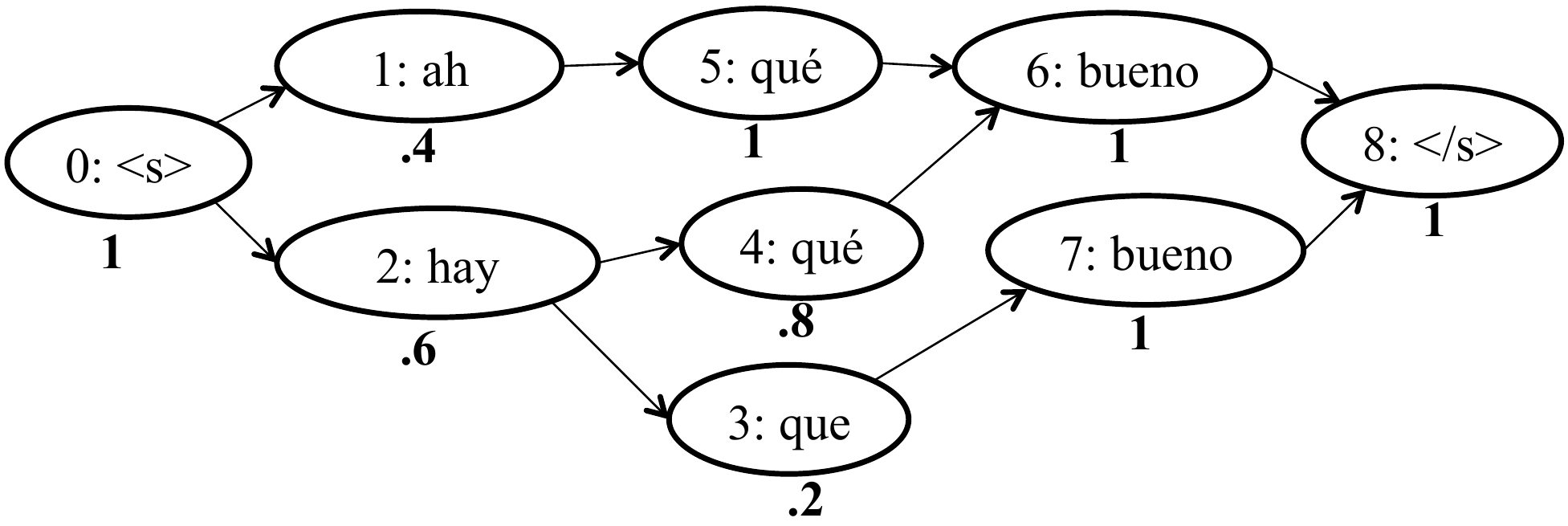}
\caption{A lattice with 3 possible paths and posterior scores. Translating the whole lattice potentially allows for recovering from errors in its 1-best hypothesis.}
\label{fig:lattice_example}
\end{figure}

Previous research on traditional phrase-based or tree-based statistical machine translation have used word lattices (e.g.\ Figure~\ref{fig:lattice_example}) as an effective tool to pass on uncertainties from a previous step \cite{Ney1999,Casacuberta2004}. Several works have shown quality improvements by translating lattices, compared to translating only the single best upstream output. Examples include translating lattice representations of ASR output \cite{Saleem2004,Zhang2005,Matusov2008}, multiple word segmentations, and morphological alternatives \cite{Dyer2008}.

Recently, neural sequence-to-sequence (seq2seq) models \cite{Kalchbrenner2013,Sutskever2014,Bahdanau2014} have often been preferred over the traditional methods for their strong empirical results and appealing end-to-end modeling.
These models force us to rethink approaches to handling lattices, because their recurrent design no longer allows for efficient lattice decoding using dynamic programming as was used in the earlier works.

As a remedy, \newcite{Su2017} proposed replacing the sequential encoder by a lattice encoder to obtain a lattice-to-sequence (lat2seq) model. This is achieved by extending the encoder's Gated Recurrent Units (GRUs) \cite{Cho2014} to be conditioned on multiple predecessor paths. The authors demonstrate improvements in Chinese-to-English translation by translating lattices that combine the output of multiple word segmenters, rather than a single segmented sequence.

However, this model does not address one aspect of lattices that we argue is critical to obtaining good translation results: their ability to encode the certainty or uncertainty of the paths through the use of posterior scores.
Specifically, we postulate that these scores are essential for tasks that require handling lattices with a considerable amount of erroneous content, such as those produced by ASR systems.
In this paper, we propose a lattice-to-sequence model that accounts for this uncertainty.
Specifically, our contributions are as follows:

\begin{itemize}[leftmargin=*]
  \item We employ the popular child-sum TreeLSTM \cite{Tai2015} to derive a lattice encoder that replaces the sequential encoder in an attentional encoder-decoder model. We show empirically that this approach yields only minor improvements compared to a baseline fine-tuned on sequential ASR outputs. This finding stands in contrast to the positive results by \newcite{Su2017}, and by \newcite{Ladhak2016} on a lattice classification task, and suggests higher learning complexity of our speech translation task.
  \item We hypothesize that lattice scores are crucial in aiding training and inference, and propose several techniques for integrating lattice scores into the model: (1) We compute {\it weighted child-sums},\footnote{This is reminiscent of the weighted pooling strategy by \newcite{Ladhak2016} for spoken utterance classification.} where hidden units in the lattice encoder are conditioned on their predecessor hidden units such that predecessors with low probability are less influential on the current hidden state. (2) We bias the TreeLSTM's {\it forget gates} for each incoming connection toward being more forgetful for predecessors with low probability, such that their cell states become relatively less influential. (3) We {\it bias the attention mechanism} to put more focus on source embeddings belonging to nodes with high lattice scores. We demonstrate empirically that the third proposed technique is particularly effective and produces strong gains over the baseline. According to our knowledge, this is the first attempt of integrating lattice scores already at the {\it training stage} of a machine translation model.
  \item We exploit the fact that our lattice encoder is a strict generalization of a sequential encoder by {\it pre-training} on sequential data, obtaining faster and better training convergence on large corpora of parallel sequential data.
\end{itemize}

We conduct experiments on the Fisher and Callhome Spanish--English Speech Translation Corpus \cite{Post2013} and report improvements of 1.4 BLEU points on Fisher and 0.8 BLEU points on Callhome, compared to a strong baseline optimized for translating 1-best ASR outputs. We find that the proposed integration of lattice scores is crucial for achieving these improvements.

\section{Background}
Our work extends the seminal work on attentional encoder-decoder models \cite{Kalchbrenner2013,Sutskever2014,Bahdanau2014} which we survey in this section. 

Given an input sequence $\boldsymbol{x}=(x_1,x_2,\dots,x_N)$, the goal is to generate an appropriate output sequence $\boldsymbol{y}=(y_1,y_2,\dots,y_M)$. The conditional probability $p(\boldsymbol{y}\mid\boldsymbol{x})$ is estimated using parameters trained on a parallel corpus, e.g.\ of sentences in the source and target language in a translation task. This probability is factorized as the product of conditional probabilities of each token to be generated: 
$p(\boldsymbol{y}\mid\boldsymbol{x})=\prod_{t=1}^{M}p(y_t\mid\boldsymbol{y}_{<t},\boldsymbol{x})$.
The training objective is to estimate parameters $\theta$ that maximize the log-likelihood of the sentence pairs in a given parallel training set $D$:
$J(\theta)=\sum_{(\boldsymbol{x},\boldsymbol{y})\in D}\log p(\boldsymbol{y}\mid\boldsymbol{x};\theta).$

\subsection{Encoder}
In our baseline model, the encoder is a bi-directional recurrent neural network (RNN), following \cite{Bahdanau2014}. Here, the source sentence is processed in both the forward and backward directions with two separate RNNs. For every input $x_i$, two hidden states are generated as
\begin{align}
\overrightarrow{\mathbf{h}}_i &=\lstm\big(E_\textit{fwd}(x_i),\overrightarrow{\mathbf{h}}_{i-1}\big) \label{form:seq_encoder}\\
\overleftarrow{\mathbf{h}}_i  &=\lstm\big(E_\textit{bwd}(x_i),\overleftarrow{\mathbf{h}}_{i+1}\big), \label{form:seq_encoder_bwd}
\end{align}
where $E_\textit{fwd}$ and $E_\textit{bwd}$ are source embedding lookup tables. We opt for long short-term memory (LSTM) \cite{Hochreiter1997} recurrent units because of their high performance and in order to later take advantage of the TreeLSTM extension \cite{Tai2015}. We stack multiple LSTM layers and concatenate the final layer into the final source hidden state $\mathbf{h}_i=\overrightarrow{\mathbf{h}}_i\mid\overleftarrow{\mathbf{h}}_i$, where layer indices are omitted for simplicity.

\subsection{Attention}
We use an attention mechanism \cite{Luong2015} to summarize the encoder outputs into a fixed-size representation. At each decoding time step $j$, a context vector $\mathbf{c}_j$ is computed as a weighted average of the source hidden states:
$\mathbf{c}_j=\sum_{i=1}^N\alpha_{ij}\mathbf{h}_i$.
The normalized attentional weights $\alpha_{ij}$ measure the relative importance of the source words for the current decoding step and are computed as a softmax with normalization factor $Z$ summing over $i$:
\begin{align}
\alpha_{ij}=\frac{1}{Z}\exp\big(\score\big(\mathbf{s}_{j-1},\mathbf{h}_i\big)\big)
\end{align}
$\score(\cdot)$ is a feed-forward neural network with a single layer that estimates the importance of source hidden state $\mathbf{h}_i$ for producing the next target symbol $y_j$, conditioned on the previous decoder state $\mathbf{s}_{j-1}$.

\subsection{Decoder}
The decoder creates output symbols one by one, conditioned on the encoder states via the attention mechanism. It contains another LSTM, initialized using the final encoder hidden state: $\mathbf{s}_0=\mathbf{h}_N$. The decoding at step $j$ assumes a special start-of-sequence symbol $y_0$ and is computed as 
$\mathbf{s}_j=\lstm\big(E_\textit{trg}(y_{j-1}),\mathbf{s}_{j-1}\big)$, and then
$\tilde{\mathbf{s}}_t=\tanh(W_\textit{hs}[\mathbf{s}_j;\mathbf{c}_j]+\mathbf{b}_\textit{hs})$
The conditional probability that the $j$-th target word is generated is:
$p(y_j\mid\boldsymbol{y}_{<j},\boldsymbol{x})=\softmax(W_\textit{so}\tilde{\mathbf{s}}_t+\mathbf{b}_\textit{so})$.
Here, $E_\textit{trg}$ is the target embedding lookup table, $W_\textit{hs}$ and $W_\textit{so}$ are weight matrices, and $\mathbf{b}_\textit{hs}$ and $\mathbf{b}_\textit{so}$ are bias vectors.

During decoding beam search is used to find an output sequence with high conditional probability.

\section{Attentional Lattice-to-Sequence Model}

The seq2seq model described above assumes sequential inputs and is therefore limited to taking a single output of an up-stream model as input. Instead, we wish to consume lattices to carry over uncertainties from an up-stream model.

\subsection{Lattices}
Lattices (e.g. Figure~\ref{fig:lattice_example}) represent multiple ambiguous or competing sequences in a compact form.
They are a more efficient alternative to enumerating all hypotheses as an $n$-best list, as they allow for avoiding redundant computation over subsequences shared between multiple hypotheses. Lattices can either be produced directly, e.g.\ by an ASR dumping its pruned search space \cite{Post2013}, or can be obtained by merging several $n$-best sequences \cite{Dyer2008,Su2017}.

A word lattice $\mathcal{G}=\langle V,E\rangle$ is a directed, connected, and acyclic graph with nodes $V$ and edges $E$. $V{\subset}\mathbb{N}$ is a node set, and $(k,i){\in}E$ denotes an edge connecting node $k$ to node $i$. $C(i)$ denotes the set of predecessor nodes for node $i$. We assume that all nodes follow a topological ordering, such that $k{<}i\ \forall\ k{\in}C(i)$. Each node $i$ is assigned a word label $w(i)$.
\footnote{It is perhaps more common to think of each edge representing a word, but we will motivate why we instead assign word labels to nodes in $\mathsection$\ref{sec:node_labeled}.}
Every lattice contains exactly one start-of-sequence node with only outgoing edges, and exactly one end-of-sequence node with only incoming edges.

\subsection{Lattices and the TreeLSTM}

One thing to notice here is that lattice nodes can have multiple predecessor states.
In contrast, hidden states in LSTMs and other sequential RNNs are conditioned on only one predecessor state ($\tilde{h}_{j}$ in left column of Table~\ref{tab:lstm}), rendering standard RNNs unsuitable for the modeling of lattices.
Luckily \newcite{Tai2015}'s TreeLSTM, which was designed to compose encodings in trees, is also straightforward to apply to lattices; the TreeLSTM composes multiple child states into a parent state, which can also be applied to lattices to compose multiple predecessor states into a successor state.
Table~\ref{tab:lstm}, middle column, shows the TreeLSTM in its child-sum variant that supports an arbitrary number of predecessors.
Conditioning on multiple predecessor hidden states is achieved by simply taking their sum as $\tilde{\mathbf{h}}_{i}$. Cell states from multiple predecessor are each passed through their own forget gates $\mathbf{f}_{jk}$ and then summed.

Encoding a lattice results in one hidden state for each lattice node. Our lat2seq framework uses this network as encoder, computing the attention over all lattice nodes.\footnote{This is similar in spirit to \newcite{Eriguchi2016} who used the TreeLSTM in an attentional tree-to-sequence model.} In other words we replace (\ref{form:seq_encoder}) by the following:

\begin{align}
\overrightarrow{\mathbf{h}}_i=\latticelstm\big(\mathbf{x}_i,\{\overrightarrow{\mathbf{h}}_{k}\mid k{\in}C(i)\}\big)
\end{align}

\begin{table*}
\centering
\renewcommand{\arraystretch}{1.3}
\begin{tabular}{| c| c| c| c|}
\hline 
    & Sequential LSTM & TreeLSTM & Proposed LatticeLSTM\tabularnewline
\hline 
\hline
recurrence
    & $\tilde{\mathbf{h}}_{i}=\mathbf{h}_{i-1}$ 
    & $\tilde{\mathbf{h}}_{i}=\sum_{k\in C(i)}\mathbf{h}_{k}$ 
    & \refstepcounter{equation}
        $\tilde{\mathbf{h}}_{i}=\sum_{k\in C(i)}\frac{w_{\textit{b/f},k}^{\mathbf{S}_\textit{h}}}{\mathbf{Z}_{\textit{h},k}}\mathbf{h}_{k}$
       (\theequation)
       \label{form:latt_lstm_childsum}
     \tabularnewline
\hline 
forget gt.
    &$\mathbf{f}_{i}=\sigma\left(W_\textit{f}\mathbf{x}_{i}+U_\textit{f}\tilde{\mathbf{h}}_{i}+\mathbf{b}_\textit{f}\right)$
    & $\begin{array} {r} \mathbf{f}_{ik} =  \sigma(W_\textit{f}\mathbf{x}_{i} + \\ U_\textit{f}\mathbf{h}_{k}+\mathbf{b}_\textit{f}) \end{array}$ 
    & \refstepcounter{equation}
        $\begin{array} {r} \mathbf{f}_{ik} =  \sigma(W_\textit{f}\mathbf{x}_{i} + U_\textit{f}\mathbf{h}_{k} + \\    \left[\ln w_{\textit{b/f},k}\mathbf{S}_\textit{f}-\mathbf{Z}_{\textit{f},k}\right]+\mathbf{b}_\textit{f}) \end{array}$ 
       \label{form:latt_lstm_forget}
        (\theequation)
        \\
\hline 
input gt.
    & \small$\mathbf{i}_{i}=\sigma\left(W_\textit{in}\mathbf{x}_{i}+U_\textit{in}\tilde{\mathbf{h}}_{i}+\mathbf{b}_\textit{in}\right)$\normalsize
    & as sequential 
    & as sequential\tabularnewline
\hline 
output gt.
    & \small$\mathbf{o}_{i}=\sigma\left(W_\textit{o}\mathbf{x}_{i}+U_\textit{o}\tilde{\mathbf{h}}_{i}+\mathbf{b}_\textit{o}\right)$\normalsize
    & as sequential 
    & as sequential\tabularnewline
\hline
update
    & \small$\mathbf{u}_{i}=\tanh\left(W_\textit{u}\mathbf{x}_{i}+U_\textit{u}\tilde{\mathbf{h}}_{i}+\mathbf{b}_\textit{u}\right)$ \normalsize
    & as sequential
    & as sequential\tabularnewline
\hline
cell
    & $\mathbf{c}_{i}=\mathbf{i}_{i}\odot \mathbf{u}_{i}+\mathbf{f}_{i}\odot \mathbf{c}_{i-1}$ 
    & $\begin{array} {l} \mathbf{c}_{i}  = \mathbf{i}_{i}\odot \mathbf{u}_{i}+ \\ \sum_{k\in C(i)}\mathbf{f}_{ik}\odot \mathbf{c}_{k} \end{array}$ 
    & as TreeLSTM\\ 
\hline
hidden
    & $\mathbf{h}_{i}=\mathbf{o}_{i}\odot \tanh(\mathbf{c}_{i})$ 
    & as sequential
    & as sequential\tabularnewline
\hline 
\hline
attention 
    & \multicolumn{2}{c|}{ $\alpha_{ij}\propto\exp\left(s\left(\cdot\right)\right)$  }
    
    & \refstepcounter{equation}
       $\alpha_{ij}{\propto}\exp\left[s\left(\cdot\right){+}S_\textit{a}\ln w_{\textit{m},i}\right]$
       (\theequation)
       \label{form:latt_attention}
       \tabularnewline
\hline 
\end{tabular}
\caption{\label{tab:lstm} Formulas for sequential and TreeLSTM encoders according to \newcite{Tai2015}, the proposed LatticeLSTM encoder, and conventional vs.\ proposed integration into the attention mechanism (bottom row). Inputs $\mathbf{x}_j$ are word embeddings or hidden states of a lower layer. $W_\cdot$ and $U_\cdot$ denote parameter matrices, $\mathbf{b}_\cdot$ bias terms, other terms are described in the text.}
\end{table*}

\begin{figure}[tb]
\includegraphics[width=7.5cm]{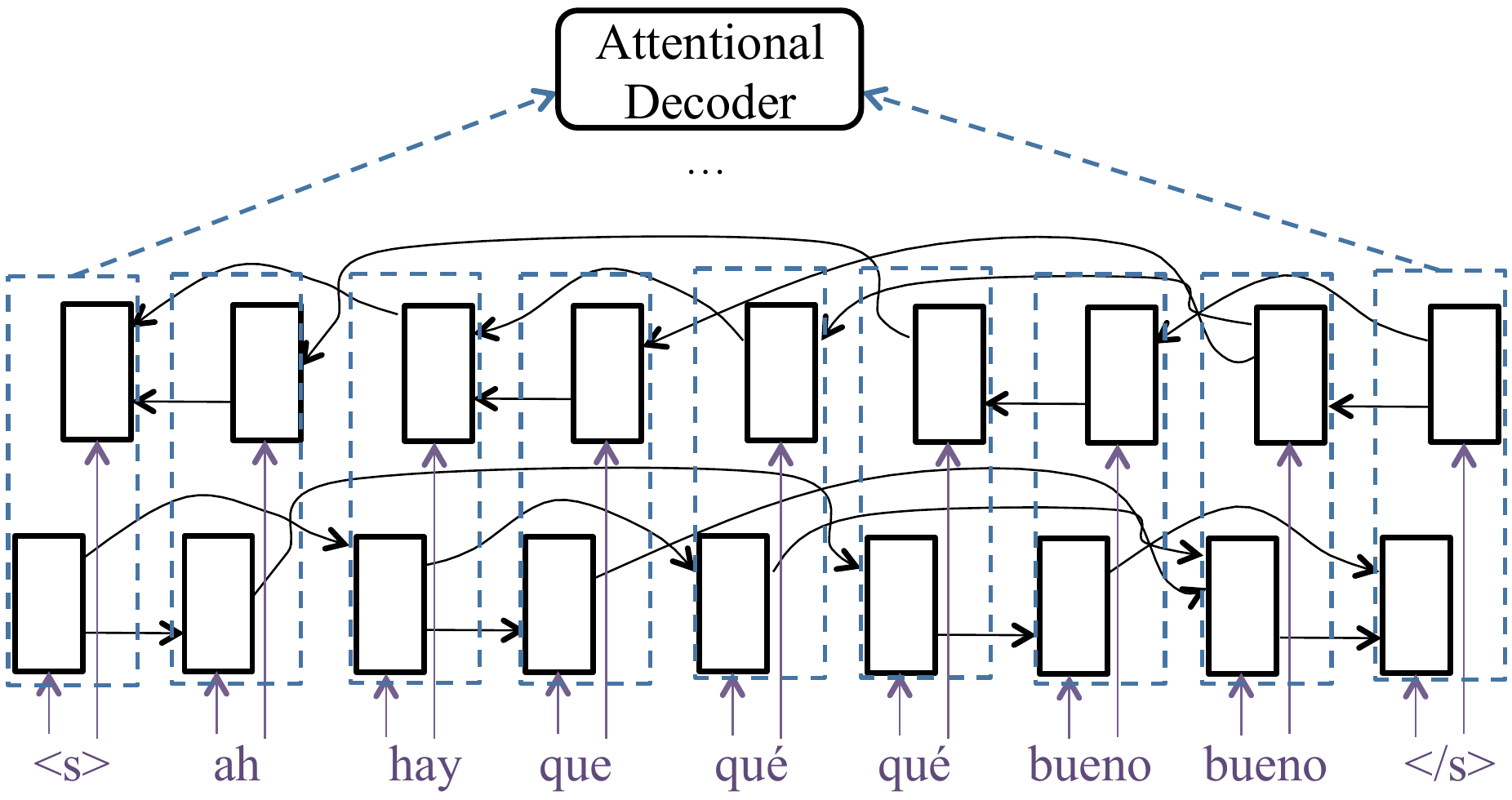}
\caption{Network structure of a bidirectional lattice encoder with one layer.}
\label{fig:lattice_lstm}
\end{figure}

Similarly, we encode the lattice in backward direction and replace (\ref{form:seq_encoder_bwd}) accordingly.
Figure~\ref{fig:lattice_lstm} illustrates the result.
The computational complexity of the encoder is $\mathcal{O}(|V|+|E|)$, i.e.\ linear in the number of nodes plus number of edges in the graph. The complexity of the attention mechanism is $\mathcal{O}(|V|M)$, where $M$ is the output sequence length. $|V|$ depends on both the expected input sentence length and the lattice density.

\subsection{Node-labeled Lattices}
\label{sec:node_labeled}
At this point we take a step back to motivate our choice of assigning word labels to lattice nodes, which is in contrast to the prior work by \newcite{Ladhak2016} and \newcite{Su2017} who assign word labels to edges. Recurrent states in edge-labeled lattice encoders are conditioned not only on multiple predecessor states, but must also aggregate words from multiple incoming edges. This implies that hidden units may represent more than one word in the lattice. Moreover, in the edge-labeled case hidden units that are in the same position in forward and backward encoders represent different words, but are nevertheless concatenated and attended to jointly. For these reasons we find our approach of encoding word-labeled lattices more intuitively appealing when used as input to an attentional decoder, although empirical justification is beyond the scope of this paper. We also note that it is easy to convert an edge-labeled lattice into a node-labeled lattice using the line-graph algorithm \cite{Hemminger1978}, which we utilize in this work.

\section{Integration of Lattice Scores}

This section describes the key technical contribution of our work: integration of lattice scores encoding input uncertainty into the lat2seq framework.
These lattice scores assign different probabilities to competing paths, and are often provided by up-stream statistical models.
For example, an ASR may attach posterior probabilities that capture acoustic evidence and linguistic plausibility of words in the lattice.
In this section, we describe our method, first explaining how we normalize scores to a format that is easily usable in our method, then presenting our methods for incorporating these scores into our encoder calculations.

\subsection{Lattice Score Normalization}

\begin{figure}[tb]
\includegraphics[width=7.5cm]{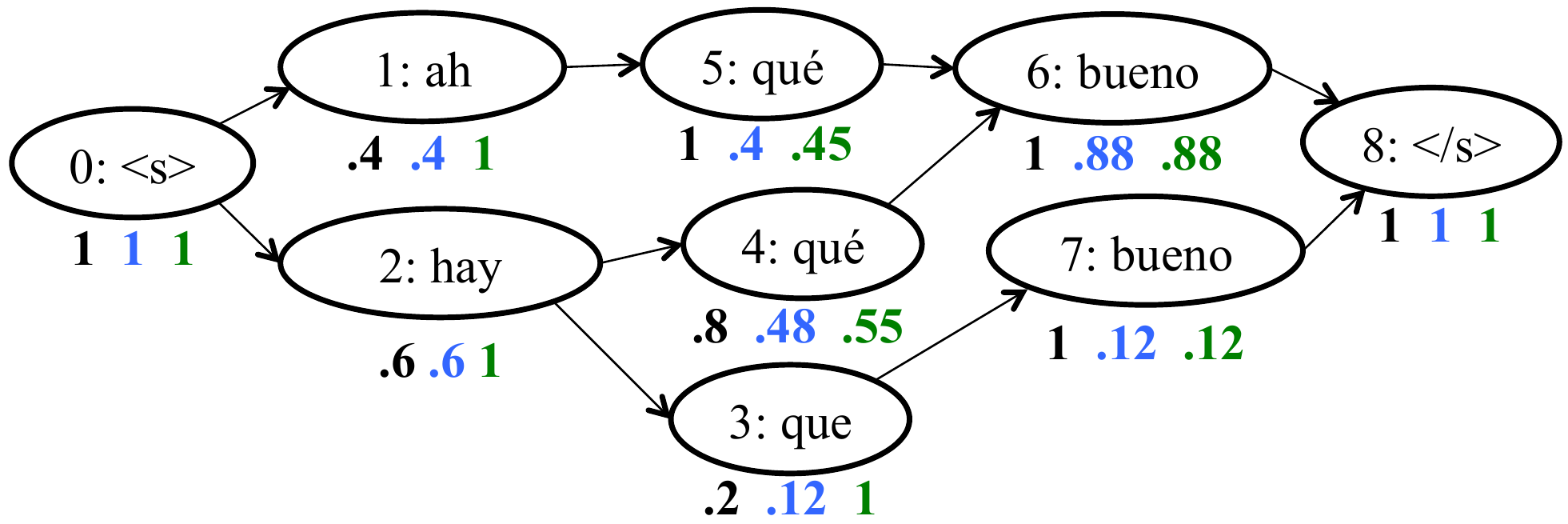}
\caption{Lattice with forward-normalized, marginal, and backward-normalized scores.}
\label{fig:lattice_scores}
\end{figure}

Lattice scores that are obtained from upstream systems (such as ASR) are typically given in forward-normalized fashion, interpreted as the probability of a node given its predecessor.
Here, outgoing edges sum up to one, as illustrated in Figure~\ref{fig:lattice_example}.
However, in some of our methods it will be necessary that scores be normalized in the backward direction, so that the weights from incoming connections sum up to one, or globally normalized, so that the probability of the node is the marginal probability of all the paths containing that node.

Let $w_{\mathit{f},i}$, $w_{\mathit{m},i}$, $w_{\mathit{b},i}$ denote forward-normalized, marginal, and backward-normalized scores for node $i$ respectively, illustrated in Figure~\ref{fig:lattice_scores}.
Given $w_{\mathit{f},i}$, we can compute marginal probabilities recursively as 
$w_{\mathit{m},i}=\sum_{k{\in}C(i)}w_{\mathit{m},k}{\cdot}w_{\mathit{f},i}$
by using the forward algorithm \cite{Rabiner1989}.
Then, we can normalize backward using 
$w_{\mathit{b},i}=\frac{w_{\mathit{m},i}}{\sum_{k{\in}C'(i)}w_{\mathit{m},k}}$,
where $C'(i)$ denotes the successors of node $i$. All 3 forms are employed in the sections below.

Furthermore, when integrating these scores into the lat2seq framework, it is desirable to maintain flexibility over how strongly they should impact the model. For this purpose, we introduce a peakiness coefficient $S$.
Given a lattice score $w_{\textit{b},i}$ in backward direction, we compute $w_{\textit{b},i}^S/Z_i$. $Z_i{=}\sum_{k{\in}C(i)}w_{\textit{b},k}$ is a re-normalization term to ensure that incoming connections still sum up to one. In the forward direction, we compute $w_{\textit{f},i}^S/Z_i$ and  normalize analogously over outgoing connections.
Setting $S{=}0$ amounts to ignoring the scores by flattening their distribution, while letting $S{\to}\infty$ puts emphasis solely on the strongest nodes.
$S$ can be optimized jointly with the other model parameters via back-propagation during model training.

\subsection{Integration Approaches}
\label{sec:lattice_scores}

We suggest three methods to integrate these scores into our lat2seq model, with equations shown in the right column of Table~\ref{tab:lstm}. These methods can optionally be combined, and we conduct an ablation study to assess the effectivity of each method in isolation ($\mathsection$\ref{sec:ablation}).

The first method consists of computing a weighted child-sum (\texttt{WCS}), using lattice scores as weights when composing the hidden state $\tilde{\mathbf{h}}_i$.
This is based on the intuition that predecessor hidden states with high lattice weights should have a higher influence on their successor than states with low weights.
The precise formulas for \texttt{WCS} are shown in (\ref{form:latt_lstm_childsum}).

The second method biases the forget gate $\mathbf{f}_{ik}$ for each predecessor cell state such that predecessors with high lattice score are more likely to pass through the forget gate (\texttt{BFG}).
The intuition for this is similar to \texttt{WCS}; the composed cell state is more highly influenced by cell states from predecessors with high lattice score.
\texttt{BFG} is implemented by introducing a bias term inside the sigmoid as in (\ref{form:latt_lstm_forget}).

In the cases of both \texttt{WCS} and \texttt{BFG}, all hidden units have their own independent peakiness.
Thus $\mathbf{S}_\textit{h}$ and $\mathbf{S}_\textit{f}$ are vectors, applied element-wise after broadcasting the lattice score.
The re-normalization terms $\mathbf{Z}_{\textit{h},k}$ and $\mathbf{Z}_{\textit{f},k}$ are also vectors and are applied element-wise.
We use backward-normalized scores $w_{\mathit{b},i}$ for the forward-directed encoder, and forward-normalized scores $w_{\mathit{f},i}$ for the backward-directed encoder.

In the third and final method, we bias the attentional weights (\texttt{BATT}) to put more focus on lattice nodes with high lattice scores.
This can potentially mitigate the problem of having multiple contradicting lattice nodes that may confuse the attentional decoder.
\texttt{BATT} is computed by introducing a bias term to the attention as in (\ref{form:latt_attention}).
Attentional weights are scalars, so here the peakiness $S_\textit{a}$ is also a scalar.
We drop the normalization term, relying instead on the softmax normalization.
Both \texttt{BFG} and \texttt{BATT} use the logarithm of lattice scores so that values will still be in the probability domain after the softmax or sigmoid is computed.

\subsection{Pre-training}

Finally, note that our lattice encoder is a strict generalization of a sequential encoder. To reduce the computational burden, we exploit this fact and perform a two-step training process where the model is first \textit{pre-trained} on sequential data, then \textit{fine-tuned} on lattice data.\footnote{For the sequential data, we set all confidence scores to 1.}
The pre-training, like standard training for neural machine translation (NMT), allows for efficient training using mini-batches, and also allows for training on standard text corpora for which we might not have lattices available.
The fine-tuning is then performed on parallel data with lattices on the source side.
This is much slower\footnote{Our implementation processed sequential inputs about 75 times faster than lattice inputs during training, and overall fine-tuning convergence was 15 times faster. Decoding was only 1.2 times slower when using lattice inputs. Note that recently proposed approaches for autobatching \cite{Neubig2017a} may considerably speed up lattice training.} than the pre-training because the network structure changes from sentence to sentence, preventing us from using efficient minibatched calculations.
However, fine-tuning for only a small number of iterations is generally sufficient, as the model is already relatively accurate in the first place.
In practice we found it important to use minibatches when fine-tuning, accumulating gradients over several examples before performing parameter updates.
This provided negligible speedups but greatly improved optimization stability.

At test time, the model is able to translate both sequential and lattice inputs and can therefore be used even in cases where no lattices are available, at potentially diminished accuracy.

\section{Experiments}

\subsection{Setting}
We conduct experiments on the Fisher and Callhome Spanish--English Speech Translation Corpus \cite{Post2013}, a corpus of Spanish telephone conversations that includes automatic transcripts and lattices. The Fisher portion consists of telephone conversations between strangers, while the Callhome portion contains telephone conversations between relatives or friends. The training data size is 138,819 sentences (Fisher/Train), and 15,000 sentences (Callhome/Train). Held-out testing data is shown in Table~\ref{tab:data}. ASR word error rates (WER) are relatively high, due to the spontaneous speaking style and challenging acoustics. Lattices contain on average 3.4 (Fisher/Train) or 4.5 (Callhome/Train) times more words than the corresponding reference transcripts.

\begin{table}[tb]
\centering
\begin{tabular}{@{}lccc@{}}
\toprule
                 & \begin{tabular}[c]{@{}c@{}}1-best\\ WER\end{tabular} & \begin{tabular}[c]{@{}c@{}}oracle\\ WER\end{tabular} & \# sent. \\ \midrule
Fisher/Dev       & 41.3                                                 & 19.3                                                 & 3,979    \\
Fisher/Dev2      & 40.0                                                 & 19.4                                                 & 3,961    \\
Fisher/Test      & 36.5                                                 & 16.1                                                 & 3,641    \\
Callhome/Devtest & 64.7                                                 & 36.4                                                 & 3,966    \\
Callhome/Evltest & 65.3                                                 & 37.9                                                 & 1,829    \\ \bottomrule
\end{tabular}
\caption{Development data statistics. Average sentence length is between 11.8 and 13.1.}
\label{tab:data}
\end{table}

For preprocessing, we tokenized and lowercased source and target sides. We removed punctuation from the reference transcripts on the source side for consistency with the automatic transcripts and lattices. All models are pre-trained and fine-tuned on Fisher/Train unless otherwise noted. Our source-side vocabulary contains all words from the automatic transcripts for Fisher/Train, replacing singletons by an unknown word token, totaling 14,648 words. Similarly, on the target side we used all words from the reference translations of Fisher/Train, replacing singletons by the unknown word, yielding 10,800 words in total.

Our implementation is based on lamtram \cite{neubig15lamtram} and the DyNet \cite{Neubig2017} toolkit. We use the implemented attentional model with default parameters: a layer size of 256 per encoder direction and 512 for the decoder. Word embedding size was also set to 512. We used two encoder layers and two decoder layers for better baseline performance. For the sequential baselines, the LSTM variant in the left column of Table~\ref{tab:lstm} was employed. We initialized the forget gate biases to 1 as recommended by \newcite{Jozefowicz2015}.

We used Adam \cite{Kingma2014} for training, with an empirically determined initial learning rate of 0.001 for pre-training and 0.0001 for fine-tuning. We halve the learning rate when the dev perplexity (on Fisher/Dev) gets worse. Pre-training and fine-tuning on 1-best sequences is performed until convergence, and training on lattices is performed for 2 epochs to keep experimental effort manageable. On Fisher/Train, this took 3-4 days on a fast CPU.\footnote{For comparison, we tried training on lattices from scratch, which took 9 days (6 epochs) to converge at a dev perplexity that was 10\% worse than with the pre-training plus fine-tuning strategy. We also confirmed BLEU scores to be much inferior without pretraining.} Minibatch size was 1000 target words for pre-training, and 20 sentences for lattice training.
Unless otherwise noted, we employed all three proposed lattice score integration approaches, and optimized peakiness coefficients jointly during training. We repeat training 3 times with different random seeds for parameter initialization and data shuffling, and report averaged results. We set the decoding beam size to 5.

\subsection{Main Results}
\label{sec:main_results}

\begin{table*}[tb]
\renewcommand{\arraystretch}{1.3}
\centering
\begin{tabular}{@{}lccclcccc@{}}
\midrule
test-time & \multicolumn{4}{c|}{Trained on}                         & \multicolumn{4}{c}{Trained on}         \\
inputs    & R    & R+1  & R+L  & \multicolumn{1}{c|}{R+L+S}         & R       & R+1      & R+L     & R+L+S   \\ 
\toprule
reference & 53.9 \textit{(7.1)} & 53.8 \textit{(6.5)} & 53.7 \textit{(6.8)} & \multicolumn{1}{c|}{54.0 \textit{(6.7)}}          & 52.2    & 51.8    & 52.2    & 52.7    \\
oracle    & 44.9 \textit{(13.4)} & 45.6 \textit{(9.5)} & 45.2 \textit{(10.6)} & \multicolumn{1}{c|}{45.2 \textit{(10.6)}}          & 44.4    & 44.6    & 44.6    & 44.8    \\ \midrule
1-best    & 35.8 \textit{(24.7)} & 37.1 \textit{(13.7)} & 36.2 \textit{(16.4)} & \multicolumn{1}{c|}{36.2 \textit{(16.3)}}          & 35.9    & 36.6    & 36.2    & 36.4    \\
lattice   & 25.9 \textit{(23.4)} & 25.8 \textit{(15.7)} & 36.9 \textit{(13.0)} & \multicolumn{1}{c|}{\textbf{38.5} \textit{(12.6)}} & 26.2    & 25.8    & 36.1    & \textbf{38.0}    \\ \midrule
          & \multicolumn{4}{c}{Fisher/Dev2}                   & \multicolumn{4}{c}{Fisher/Test}
\end{tabular}
\caption{BLEU scores (4 references) and perplexities \textit{(in brackets)}. Models are pre-trained only (R), fine-tuned on either 1-best outputs (R+1), lattices without scores (R+L), or lattices with scores (R+L+S). Statistically significant improvement (paired bootstrap resampling, $p<0.05$) over 1-best/R+1 is in bold.}
\label{tab:main_results}
\end{table*}

We compare 4 systems: Performing pre-training on the sequential reference transcripts only (R), fine-tuning on 1-best transcripts (R+1), fine-tuning on lattices without scores (R+L), and fine-tuning on lattices including lattice scores (R+L+S). At test time, we try references, lattice oracles,\footnote{The path through the lattice with the best WER.} 1-best transcripts, and lattices as inputs to all 4 systems. The former 2 experiments give upper bounds on achievable translation accuracy, while the latter 2 correspond to a realistic setting. Table~\ref{tab:main_results} shows the results on Fisher/Dev2 and Fisher/Test.

Before even considering lattices, we can see that 1-best fine-tuning boosted BLEU scores quite impressively (1-best/R vs.\ 1-best/R+1), with gains of 1.3 and 0.7 BLEU points. This stands in contrast to \newcite{Post2013} who find the 1-best transcripts not to be helpful for training a hierarchical machine translation system. Possible explanations are learning from repeating error patterns, and improved robustness to erroneous inputs. On top of these gains, our proposed set-up (lattice/R+L+S) improve BLEU scores by another 1.4. Removing the lattice scores (lattice/R+L) diminishes the results and performs worse than the 1-best baseline (1-best/R+1), indicating that the proposed lattice score integration is crucial for good performance. This demonstrates a clear advantage of our proposed method over that of \newcite{Su2017}.

As can be seen in the table, models fine-tuned on lattices show reasonable performance for both lattice and sequential inputs (1-best/R+L, lattice/R+L, 1-best/R+L+S, lattice/R+L+S). This is not surprising, given that the lattice training data includes lattices of varying density, including lattices with very few paths or even only one path. On the other hand, without fine-tuning on lattices, using lattices as input performs poorly (lattice/R and lattice/R+1). A closer look revealed that translations were often too long, potentially because implicitly learned mechanisms for length control were not ready to handle lattice inputs.

Table~\ref{tab:main_results} reports perplexities for Fisher/Dev2. Unlike the corresponding BLEU scores, the lattice encoder appears stronger than the 1-best baseline in terms of perplexity even without lattice scores (lattice/R+L vs.\ 1-best/R+1). To understand this better, we computed the average entropy of the decoder softmax, a measure of how much confusion there is in the decoder predictions, independent of whether it selects the correct answer or not. Over the first 100 sentences, this value was 2.24 for 1-best/R+1, 2.39 for lattice/R+L, and 2.15 for lattice/R+L+S. This indicates that the decoder is more confused for lattices without scores, while integrating lattice scores removes this problem. These numbers also suggest that it may be possible to obtain further gains using methods that stabilize the decoder.

\subsection{Ablation Experiments}
\label{sec:ablation}
Next, we conduct an ablation study to assess the impact of the three proposed extensions for integrating lattice scores ($\mathsection$\ref{sec:lattice_scores}). We train models with different peakiness coefficients $S$, either ignoring lattices scores by fixing $S{=}0$, using lattice scores as-is by fixing $S{=}1$, or optimizing $S$ during training. Table~\ref{tab:ablation} shows the results. Overall, joint training of $S$ gives similar results as fixing $S{=}1$, but both clearly outperform fixing $S{=}0$. Removing confidences (setting $S{=}0$) in one place at a time reveals that the attention mechanism is clearly the most important point of integration, while gains from the integration into child-sum and forget gate are smaller and not always consistent.

We also analyzed what  peakiness values were actually learned. We found that all 3 models that we trained for the averaging purposes converged to $S_\mathit{a}{=}0.62$. $\mathbf{S}_\mathit{h}$ and $\mathbf{S}_\mathit{f}$ had per-vector means between 0.92 and 1.0, at standard deviations between 0.02 and 0.04. We conclude that while the peakiness coefficients were not particularly helpful in our experiments, stable convergence behavior makes them safe to use, and they might be helpful on other data sets that may contain lattice scores of higher or lower reliability.

\begin{table}[tb]
\centering
\begin{tabular}{ccccc}
\hline
\begin{tabular}[c]{@{}c@{}} \texttt{BATT} \\ $S_a$\end{tabular}     & \begin{tabular}[c]{@{}c@{}} \texttt{WCS} \\ $\mathbf{S}_h$\end{tabular}     & \begin{tabular}[c]{@{}c@{}} \texttt{BFG} \\ $\mathbf{S}_f$\end{tabular}     & \begin{tabular}[c]{@{}l@{}}Fisher\\ /Dev2\end{tabular} & \begin{tabular}[c]{@{}l@{}}Fisher\\ /Test\end{tabular} \\ \hline
0        & $\mathbf{0}$        & $\mathbf{0}$        & 36.9                                                  & 36.1                                                       \\
1        & $\mathbf{1}$        & $\mathbf{1}$        & \textbf{38.2}                                     & \textbf{37.4}                                                        \\
*        & *                            & *                           & \textbf{38.5}                                     & \textbf{38.0}                                                       \\
\hline
0        & $\mathbf{1}$        & $\mathbf{1}$        & 37.2                                                  & 36.2                                                       \\
1        & $\mathbf{0}$        & $\mathbf{1}$        & \textbf{37.9}                                      & \textbf{37.5}                                                       \\
1        & $\mathbf{1}$        & $\mathbf{0}$        & \textbf{38.2}                                      & \textbf{37.8}                                                       \\
\hline
0        & *                           & *                           & 37.0                                                  & 36.3                                                       \\
*        & $\mathbf{0}$        & *                            & \textbf{38.3}                                   & \textbf{37.9}                                                       \\
*        & *                            & $\mathbf{0}$        & \textbf{38.1}                                   & \textbf{37.5}                                                       \\ \hline
\multicolumn{3}{l}{1-best/R+1}                      & 37.2                                                   & 36.6                                                      
\end{tabular}
\caption{BLEU scores (4 references) for differently configured peakiness coefficients $S_a, \mathbf{S}_h, \mathbf{S}_f$. 0/1 means fixing to that value, * indicates optimization during training. Statistically significant improvement over 1-best/R+1 is in bold.}
\label{tab:ablation}
\end{table}

\subsection{Callhome Experiments}
In this experiment, we test a situation in which we have a reasonable amount of sequential data available for pre-training, but only a limited amount of lattice training data for the fine-tuning step. This may be a more realistic situation, because speech translation corpora are scarce. To investigate in this scenario, we again pre-train our models on Fisher/Train, but then fine-tune them on the 9 times smaller Callhome/Train portion of the corpus. We fine-tune for 10 epochs, all other settings are as before. We use Callhome/Evltest for testing. Table~\ref{tab:callhome} shows the results. The trends are consistent to $\mathsection$\ref{sec:main_results}: The proposed model (lattice/R+L+S) outperforms the 1-best baseline (1-best/R+1) by 0.8 BLEU points, which in turn beats the pre-trained system (1-best/R) by 1.5 BLEU points. Including the lattice scores is clearly beneficial, although lattices without scores also improve over 1-best inputs in this experiment.

\begin{table}[tb]
\centering
\begin{tabular}{@{}lcccc@{}}
\toprule
test-time  & \multicolumn{4}{c}{Trained on} \\ 
inputs      & R        & R+1     & R+L     & R+L+S  \\ \midrule
reference & 24.7   & 24.3     & 24.8    & 24.4   \\
oracle      & 15.8   & 16.8      & 16.3  & 15.9   \\ \midrule
1-best      & 11.8   & 13.3     & 12.4     & 12.0   \\
lattice       & 9.3    & 7.1       & \textbf{13.7}    & \textbf{14.1}   \\ \bottomrule
\end{tabular}
\caption{BLEU scores on Callhome/Evltest (1 reference). All models are pre-trained on Fisher/Train references (R), and potentially fine-tuned on Callhome/Train. The best result using 1-best or lattice inputs is in bold. Statistically significant improvement over 1-best/R+1 is in bold.}
\label{tab:callhome}
\end{table}

\subsection{Impact of Lattice Quality}
Next, we analyze the impact of using lattices and lattice scores as the ASR WER changes. We concatenate all test data from Table~\ref{tab:data} and divide the result into bins according to the 1-best WER. We sample 1000 sentences from each bin, and compare BLEU scores between several models.

\begin{figure}[tb]
\includegraphics[width=7.5cm]{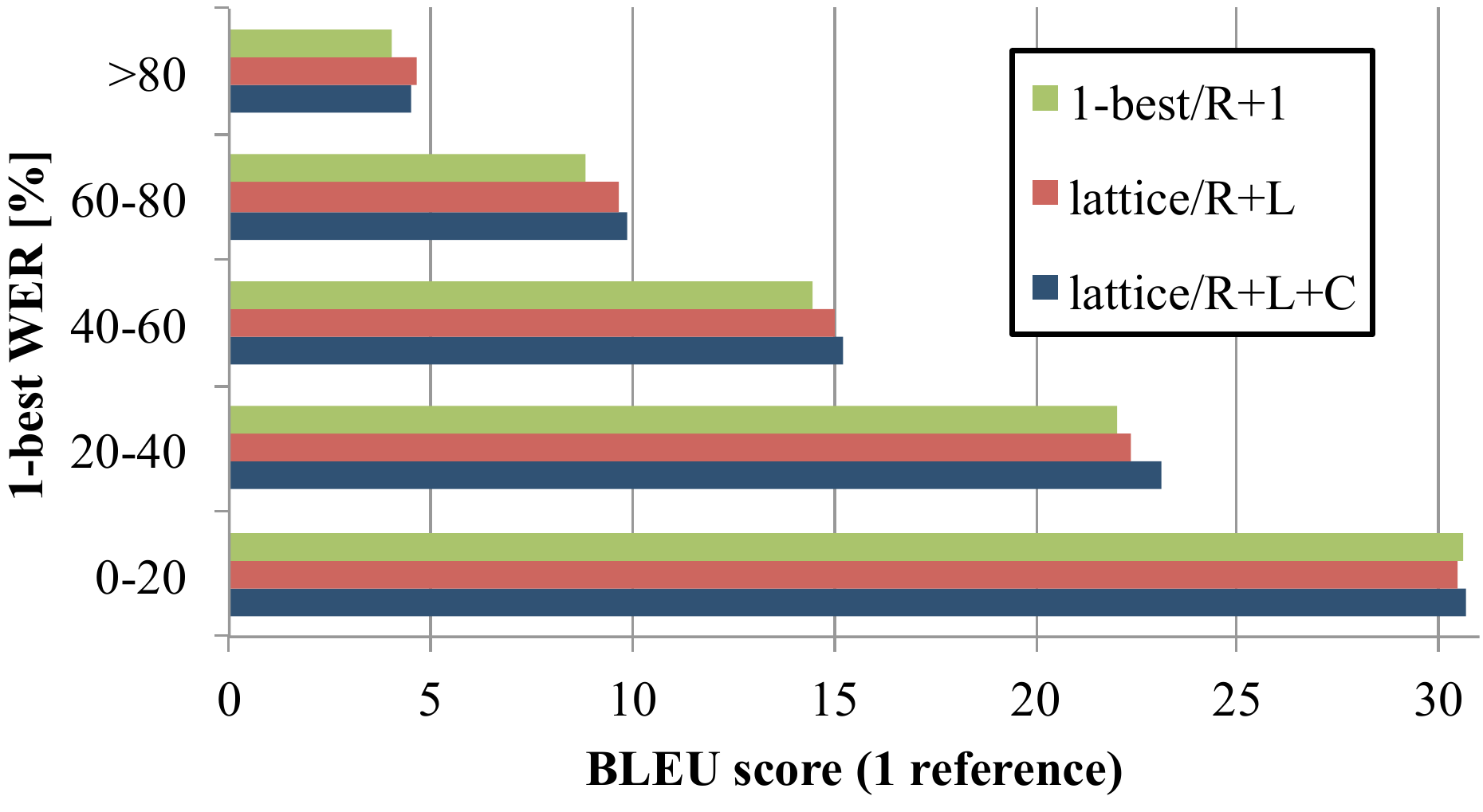}
\caption{BLEU score over varying 1-best WERs.}
\label{fig:bleu-vs-wer}
\end{figure}

The results are shown in Figure~\ref{fig:bleu-vs-wer}. For very good WERs, lattices do not improve over 1-best inputs, which is unsurprising. In all other cases, lattices are helpful. Lattice scores are most beneficial for moderate WERs, and not beneficial for very high WERs. We speculate that for high WERs, the lattice scores tend to be less reliable than for lower WERs.

\section{Conclusion}

We investigated translating uncertain inputs from an error-prone up-stream component using a neural lattice-to-sequence model. Our proposed model takes word lattices as input and is able to take advantage of lattice scores. In our experiments in a speech translation task we find consistent improvements over translating 1-best transcriptions and that consideration of lattice scores, especially in the attention mechanism, is crucial for obtaining these improvements.

Promising avenues for future work are investigating consensus networks \cite{Mangu2000a} for potential gains in terms of speed or quality as compared to lattice inputs, explicitly dealing with rare or unknown words in the lattice, and facilitating GPU training via autobatching \cite{Neubig2017a}.

\section*{Acknowledgments}

We thank Paul Michel and Austin Matthews for their helpful comments on earlier drafts of this paper.

\bibliography{library}
\bibliographystyle{emnlp_natbib}

\clearpage

\appendix

\section{Appendix: Cherry-Picked Examples}
\label{sec:supplemental}
\subsection{Erroneous 1-best}
"ideal" is contained in lattice but not 1-best transcript, and correctly translated in the lattice/R+L+S setting.
\begin{description}
 \item[1-best/R+1 input:] y y eso es algo que a mi me parece contraproducente verdad porque uno piensa y cuando ya a todos uno quisiera tal vez \textbf{un mundo} ya el de que una vez que cadena cuerpos trabajar\'{a}n por el bienestar de de todos
 \item[1-best/R+1 output:] and , and that 's something that seems to me , right ? because one thinks , and when you think , and when everyone would like perhaps \textbf{a world} already , the one time that the chain changes for the
 \item[lattice/R+L+S input:] 
\includegraphics[width=5.5cm]{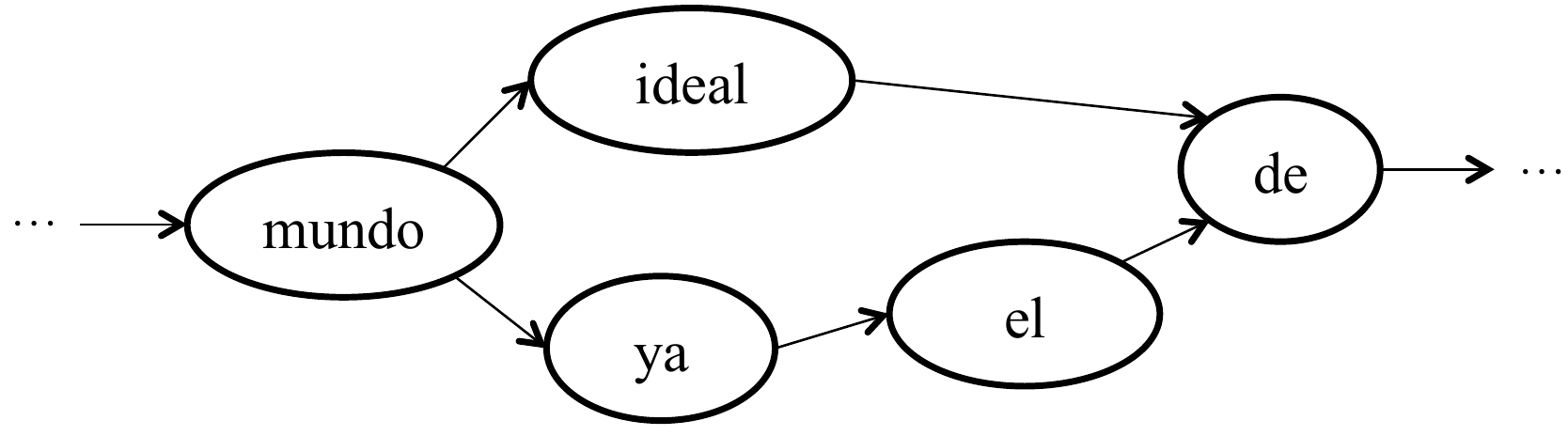}
 \item[lattice/R+L+S output:] and , and that 's something that seems to me , right ? because one thinks , and when you see , when you go to \textbf{a ideal world} , you see that they are illegals for the , well , they are all foreigners
\end{description}

\subsection{Redundant Lattice Content}
Another frequent pattern was a word appearing (once) in the 1-best transcript, but multiple times in the lattice, and only the lattice/R+L+S model translating this word.
\begin{description}
 \item[1-best/R+1 input:] los que van porque que es un d\'{i}a los que van porque no tiene alicia derrita jugar y los que s\'{i} caray \textbf{profesionalmente} porque hay ciertos counselor bueno creo que soy jos\'{e} playa que
 \item[1-best/R+1 output:] the ones that go , because it 's a day that they go , because they don 't have alicia , play and the ones that are italian , because there are some \texttt{<unk>} , well , i think i 'm jose
 \item[lattice/R+L+S input:] 
\includegraphics[width=5.5cm]{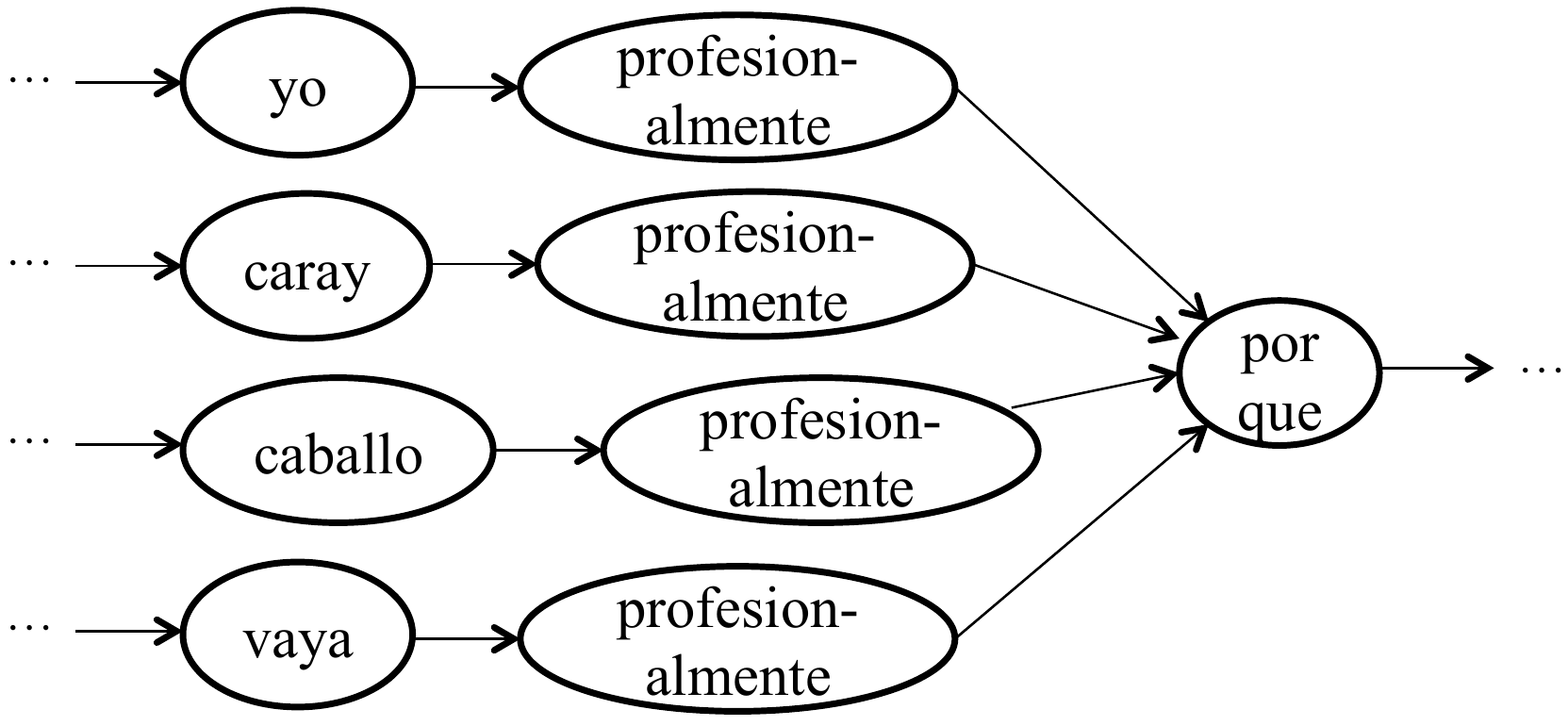}
 \item[lattice/R+L+S output:] the ones that go , because it 's a day that they go because you don 't want to play and play , and the ones that influenced \textbf{professionally} , because there are certain things , well , i think that i 'm jose
\end{description}

\subsection{Redundant Lattice Content}
In this example, the lattice/R+L+S system correctly produced \texttt{<unk>} as translation to the unknown input word "m\'{o}viles", while the 1-best/R+1 system produced a seemingly random word instead. A possible explanation is the added context helping the lat2seq system to know when to be unsure.
\begin{description}
 \item[1-best/R+1 input:] s\'{i} s\'{i} bueno contar otro que usaban los tel\'{e}fonos sat\'{e}lite los los \textbf{m\'{o}viles} no funcionaban bien porque pero a veces si funcionaban s\'{i}
 \item[1-best/R+1 output:] yes , well , tell me that i used to use satellite phones , the \textbf{kids} didn 't work well , because sometimes it worked , yes
 \item[lattice/R+L+S input:] 
\includegraphics[width=5.5cm]{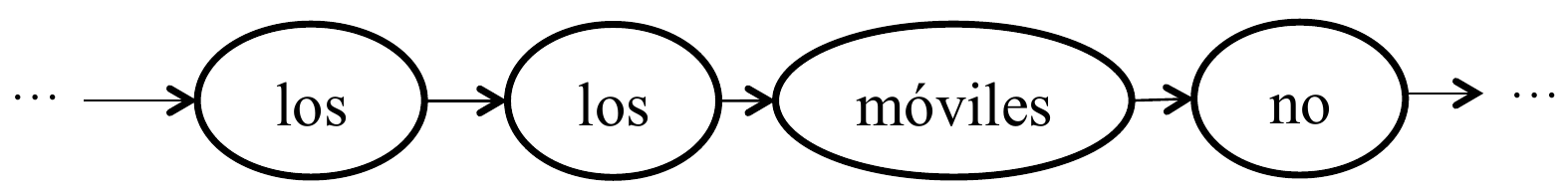}
 \item[lattice/R+L+S output:] yes , yes , well , with the other that i used to use the satellite phones , the , the $\langle$\textbf{unk}$\rangle$ didn 't work well because , but sometimes it worked , yes
\end{description}

\subsection{Counter Example}
In this example, the lattice (but not the 1-best transcript) contained then word "san", which tricked the lat2seq decoder to produce "san francisco".
\begin{description}
 \item[1-best/R+1 input:] pero los dem\'{a}s aqu\'{i} est\'{a}n tambi\'{e}n s\'{i} est\'{a} bien est\'{a} tranquilo para ac\'{a}
 \item[1-best/R+1 output:] but the rest here are also , yes , it 's ok , it 's quiet for here
 \item[lattice/R+L+S input:] 
\includegraphics[width=5.5cm]{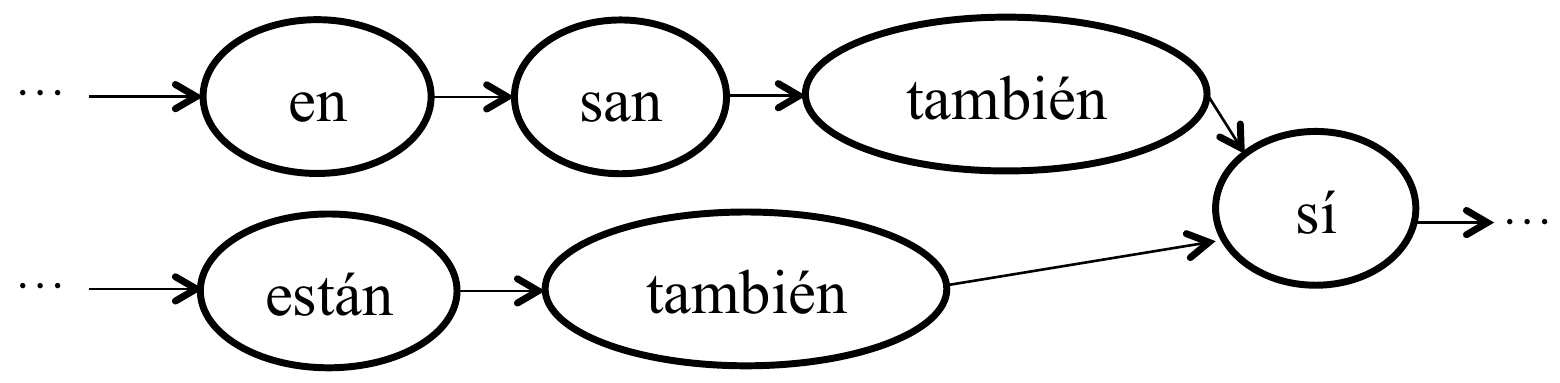}
 \item[lattice/R+L+S output:] but the rest here in \textbf{san francisco} is very quiet here
\end{description}

\end{document}